\documentclass[10pt,twocolumn,letterpaper]{article}

\usepackage{iccv}
\usepackage{times}
\usepackage{epsfig}
\usepackage{graphicx}
\usepackage{amsmath}
\usepackage{amssymb}

\usepackage[sort&compress,numbers]{natbib}
\usepackage{booktabs}
\usepackage{multirow}
\usepackage{caption}
\usepackage{color}
\usepackage{colortbl}
\usepackage[dvipsnames,svgnames,x11names]{xcolor} 
\definecolor{mygray}{gray}{.9}
\definecolor{light-gray}{gray}{0.5}
\definecolor{pretty-blue}{RGB}{0, 113, 188}
\definecolor{linecolor}{gray}{.945} 
\definecolor{linecolor1}{gray}{.895} 
\definecolor{kaiming-green}{RGB}{57,181,74} 
\definecolor{icmlblue}{rgb}{0,0.08,0.45} 
\definecolor{sun-orange}{RGB}{238,133,62} 
\definecolor{my_red}{RGB}{214, 39, 40}

\usepackage{amssymb}
\usepackage{pifont}
\newcommand{\cmark}{\ding{51}}%
\newcommand{\xmark}{\ding{55}}%

\usepackage[colorlinks=True,
            linkcolor=red,
            anchorcolor=blue,  
            pagebackref,
            citecolor=pretty-blue,
            breaklinks=True,
            ]{hyperref}

\iccvfinalcopy 

\def\iccvPaperID{2680} 

\ificcvfinal\fi

\begin{document}

\title{CLIP-FO3D: Learning \underline{F}ree \underline{O}pen-world \underline{3D} Scene Representations \\ from 2D Dense CLIP}

\author{Junbo Zhang\\
Tsinghua University\\
{\tt\small zhangjb21@mails.tsinghua.edu.cn}
\and
Runpei Dong\\
Xi'an Jiaotong University\\
{\tt\small runpei.dong@stu.xjtu.edu.cn}
\and
Kaisheng Ma\thanks{Corresponding Author.}\\
Tsinghua University\\
{\tt\small kaisheng@mail.tsinghua.edu.cn}
}

\maketitle
\ificcvfinal\thispagestyle{empty}\fi

\begin{abstract}
\vspace{-0.3cm}
   Training a 3D scene understanding model requires complicated human annotations, which are laborious to collect and result in a model only encoding close-set object semantics. In contrast, vision-language pre-training models (e.g., CLIP) have shown remarkable open-world reasoning properties. To this end, we propose directly transferring CLIP's feature space to 3D scene understanding model without any form of supervision. We first modify CLIP's input and forwarding process so that it can be adapted to extract dense pixel features for 3D scene contents. We then project multi-view image features to the point cloud and train a 3D scene understanding model with feature distillation. Without any annotations or additional training, our model achieves promising annotation-free semantic segmentation results on open-vocabulary semantics and long-tailed concepts. Besides, serving as a cross-modal pre-training framework, our method can be used to improve data efficiency during fine-tuning. Our model outperforms previous SOTA methods in various zero-shot and data-efficient learning benchmarks. Most importantly, our model successfully inherits CLIP's rich-structured knowledge, allowing 3D scene understanding models to recognize not only object concepts but also open-world semantics.
\end{abstract}

\vspace{-0.5cm}

\section{Introduction}
3D scene understanding aims to distinguish objects' semantics, identify their locations, and infer the geometric attributes from 3D scene data. It has a wide range of applications in virtual reality~\cite{Park2008Multiple3O}, robot navigation~\cite{Ahn2022DoAI,Oh2002DevelopmentOS} and autonomous driving~\cite{Lang2019PointPillars,Yin2021CenterPoint}. However, training traditional 3D scene understanding models requires a large number of human annotations, which are laborious to collect. Besides, the human annotations used in the current 3D scene understanding benchmark only contain close-set semantic information of the objects (\eg, 20 classes in ScanNet~\cite{Dai2017ScanNetR3}). These make it difficult for 3D scene understanding systems to recognize open-vocabulary categories and to infer open-world semantics.

Large-scale vision-language foundation models (\eg, CLIP~\cite{Radford2021LearningTV}) capture rich visual and language features. They only require image-text pairs mined from the Internet for unsupervised pre-training, and have demonstrated superior ability in zero-shot and open-vocabulary reasoning for classification and dense prediction tasks~\cite{Ghiasi2021OpenVocabularyIS,Li2022LanguagedrivenSS,Zabari2021SemanticSI,Zhou2021ExtractFD}. However, due to the difficulty of collecting 3D-text pairs and the complexity of scene data, is extremely challenging to build analogous 3D foundation models. Moreover, transferring CLIP's capabilities to 3D scene understanding models is still an understudied problem.

\begin{figure}[t]
\begin{center}
\centerline{\includegraphics[width=0.90\columnwidth]{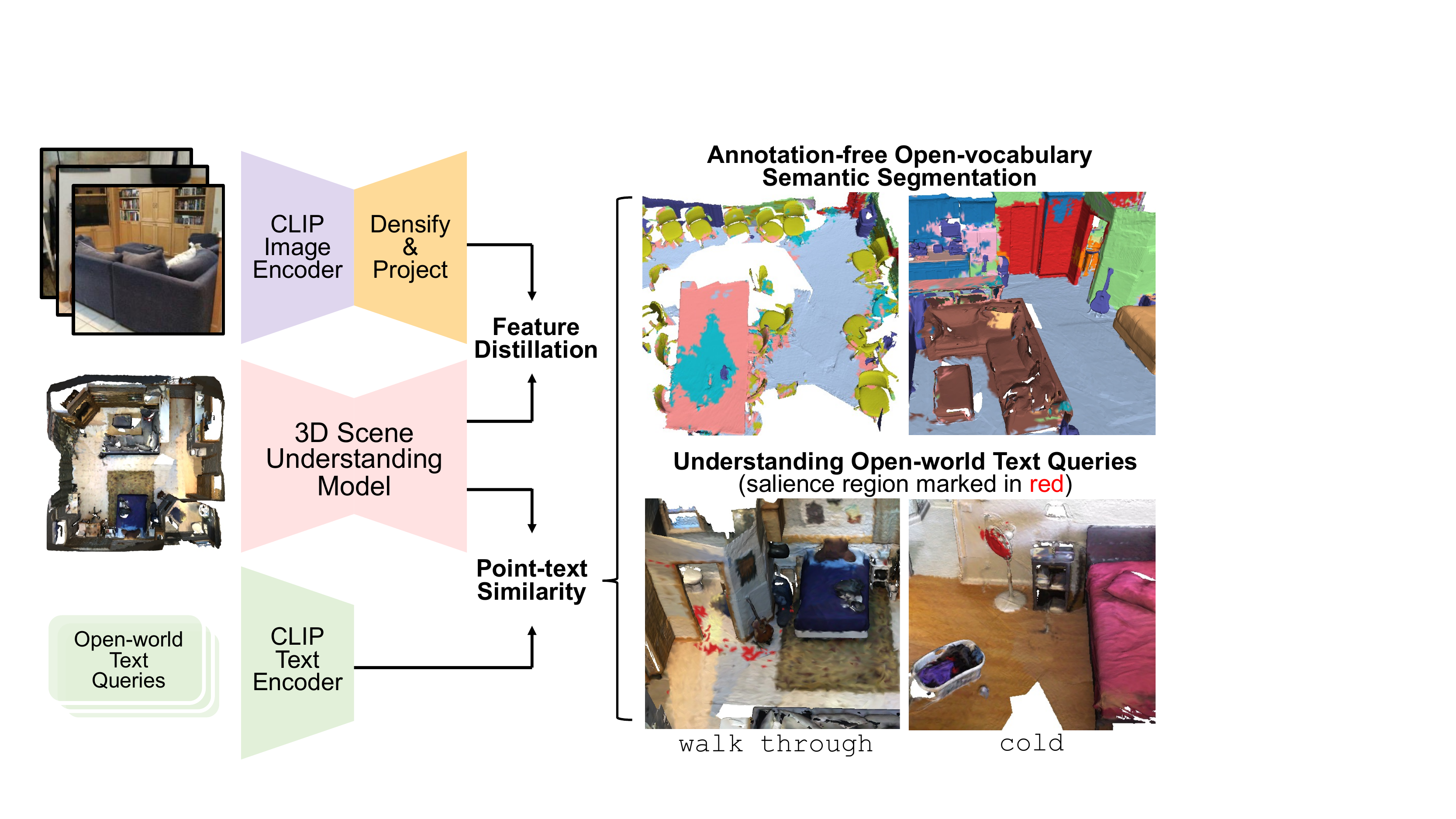}}
\caption{Overview of CLIP-FO3D. We directly transfer CLIP's feature space to 3D point representations by extracting dense pixel features, 2D-to-3D feature projection, and feature distillation. Without any human annotations, CLIP-FO3D successfully utilizes the point-text feature similarity to perform semantic segmentation on open-vocabulary semantics. Besides, CLIP-FO3D inherently encodes CLIP's open-world knowledge, broadening the applications of 3D scene understanding models.}
\label{front}
\end{center}
\vspace{-1.2cm}
\end{figure}

In this paper, we propose directly transferring CLIP's feature space to 3D scene understanding models without any supervision (\eg, 2D/3D annotations or vision-language grounding annotations). Once the transfer is complete, our model can complete the open-vocabulary 3D scene understanding task without additional annotation or training. In this way, we intend to preserve the rich information inherited from CLIP model to the maximum. It is because that training with any human annotations restricts the feature space to a limited set of vocabularies and loses CLIP's powerful open-world properties.

Recent methods in 3D vision that directly distill CLIP's open-vocabulary knowledge to 3D models focus only on the object-level data ~\cite{Huang2022CLIP2PointTC,Liu2022PartSLIPLP,Zhang2021PointCLIPPC,Zhu2022PointCLIPVA}. 
By contrast, it is much more difficult to directly transfer CLIP's feature space to 3D models on scene-level data. Because CLIP's vision encoder only extract image-level global feature, while 3D scene understanding requires dense point-level features. Although MaskCLIP~\cite{Zhou2021ExtractFD} in 2D vision has taken the first step towards extracting pixel-level features from CLIP's final feature map, it has difficult adapting to 3D scene's contents since they contain much more objects and more complex structures. Besides, it has an inherent defect in precisely locating and segmenting objects' boundaries due to the limited resolution of CLIP's final feature map.

To address the abovementioned problems, we present a new method to extract free pixel-level CLIP features without breaking CLIP's feature space. We first crop the input images at multi-scale to accommodate various object sizes. To preserve object-level semantics, we divide each cropped sample into semantically relevant regions and extract a local feature for each region. In the ViT~\cite{dosovitskiy2020vit} encoder, we add several local classification tokens, which only aggregate information from local patches within a region. By simply forwarding CLIP's encoder, we extract a pixel-level feature map of each 3D scene's RGB view.

After extracting pixel-level CLIP features, we adopt the feature projection scheme in 3DMV~\cite{Dai20183DMVJ3} to project multi-view image features to the point cloud. The resulting point features are aligned with CLIP's feature space and are used as off-line training targets. Then we train the 3D understanding model with feature distillation, minimizing the distance between learned point features and the target features. This way, we obtain CLIP-FO3D, which extracts free and open-world 3D scene representations aligned with CLIP.

Our CLIP-FO3D can perform annotation-free open-vocabulary 3D semantic segmentation without additional training processes. Since CLIP's vision features are well aligned with text features, we can take the text embeddings of each class name's prompts as classification weights to perform semantic segmentation. CLIP-FO3D performs remarkably on standard close-set ScanNet~\cite{Dai2017ScanNetR3} and S3DIS~\cite{Armeni20163DSP} segmentation benchmark. In addition, to examine CLIP-FO3D's open-vocabulary capability inherited from CLIP, we extend the label set of the standard ScanNet dataset with NYU labels~\cite{Silberman2012IndoorSA}. We demonstrate that CLIP-FO3D has remarkable segmentation results on NYU-40 classes and other long-tailed categories beyond the NYU label set.

Besides, given that collecting 3D point clouds and annotating are laborious, CLIP-FO3D can also be regarded as an unsupervised cross-modal pre-training framework to benefit data efficiency. CLIP-FO3D achieves great performance in traditional benchmarks where limited annotations are provided, such as zero-shot and data-efficient learning. 

Most importantly, CLIP-FO3D encodes rich open-world knowledge inherited from CLIP. It understands not only object concepts, but also text queries with open world semantics, broadening the application of 3D scene understanding. 

Our contributions can be summarized as follows:

\begin{itemize}
\item We propose directly transferring CLIP's feature space to 3D scene understanding models without any supervision, which preserves CLIP's open-world properties to the maximum.

\item We present a method to extract pixel-level CLIP features, and a feature distillation method to align 3D point representations with CLIP's feature space.

\item Our model achieves promising annotation-free 3D semantic segmentation performance on large vocabularies, and shows remarkable open-world properties.

\item As an unsupervised pre-training method, our model outperforms previous state-of-the-art methods in zero-shot and data-efficient learning.
\end{itemize}

\begin{figure*}[t]
\begin{center}
\centerline{\includegraphics[width=0.95\linewidth]{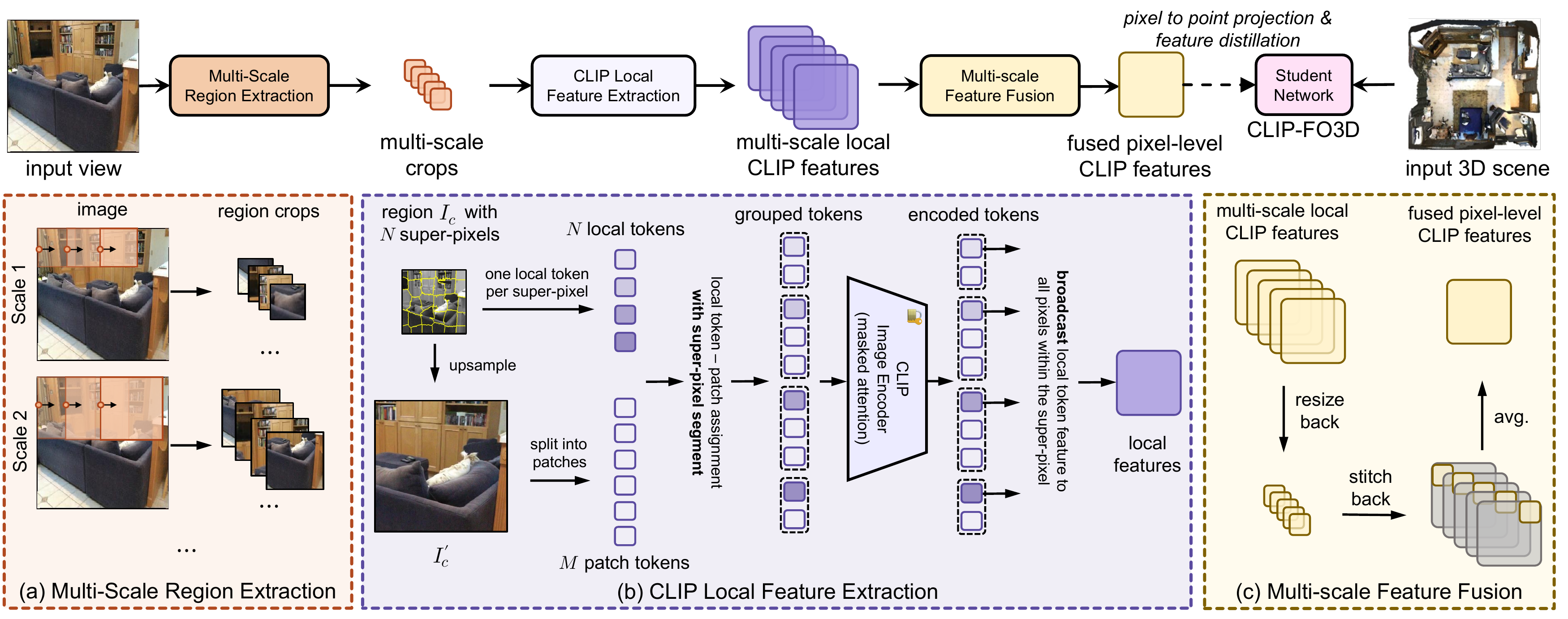}}
\vspace{-10pt}
\caption{\textbf{Top:} Overall training process for CLIP-FO3D. For RGB views, we extract free pixel-level CLIP features and project pixel features to 3D points. For 3D scenes, we encode the point features and train CLIP-FO3D with feature distillation. \textbf{Bottom:} Three steps to extract pixel-level CLIP features. \textbf{(a)} The input image is cropped at multi-scale to accommodate various object sizes.
\textbf{(b)} Extracting pixel-level feature map for each region crop. Each region crop $I_\text{c}$ is divided into $N$ super-pixels, and a local token is initialized for each super-pixel. The crop is then upsampled and input into CLIP's ViT encoder together with the local tokens. Each patch token in ViT is grouped with a local token based on the spatial location of patch and super-pixel. The attention score of local token is computed only from its assigned patches. The encoded local token feature is finally broadcast to all pixels within the super-pixel. \textbf{(c)} All region crops' feature maps are resized back and stitched to the original image. We average the multiple features of pixels to obtain final pixel-level CLIP features.}
\label{main_fig}
\end{center}
\vspace{-1.1cm}
\end{figure*}

\vspace{-0.4cm}
\section{Related Work}
\subsection{Open-vocabulary Dense Prediction in 2D Vision}
\vspace{-0.2cm}

Open-vocabulary dense prediction aims to recognize and localize objects with open-set semantics. Pioneering works~\cite{Xu2022GroupViTSS,Zareian2020OpenVocabularyOD,Zhou2022DetectingTC,LIU2022OpenworldSS} in 2D vision mainly utilize large-scale image-caption annotations as weak supervision source to enlarge the vocabulary set. Other works~\cite{Du2022LearningTP,Feng2022PromptDetTO,Ghiasi2021OpenVocabularyIS,Gu2021OpenvocabularyOD,Li2022LanguagedrivenSS,Liang2022OpenVocabularySS,Lddecke2021ImageSU,Rasheed2022BridgingTG,Zang2022OpenVocabularyDW,Zhong2021RegionCLIPRL,Mukhoti2022OpenVS} distill vision-language foundation models' (\eg, CLIP) knowledge to transfer their open-vocabulary capability. However, they all require some form of human annotations, such as box/mask proposals and pixel semantic annotations. Open-world knowledge encoded in CLIP is \textit{forgotten} when fine-tuning with human annotations. In contrast, MaskCLIP~\cite{Zhou2021ExtractFD} proposes directly utilizing CLIP for dense prediction tasks without training. However, it struggles to handle 3D scene's RGB views with more complex contents, as shown in Section \ref{experiments}. We present a new method to extract free pixel-level CLIP features from 3D scene's RGB views by only modifying CLIP's inputs and forwarding process \textit{without} fine-tuning.

\subsection{Zero-shot 3D Visual Recognition}
\vspace{-0.2cm}
Zero-shot learning is relatively under-explored in 3D. Most works focus on recognition tasks on object-level data~\cite{Cheraghian2019MitigatingTH,Cheraghian2019TransductiveZL,Cheraghian2021ZeroShotLO,Cheraghian2019ZeroshotLO,Qi2023ContrastWR,Xue2023ULIP,Hess2022LidarCLIPOH}. Recent works~\cite{Huang2022CLIP2PointTC,Zhang2021PointCLIPPC} adopt CLIP to perform zero-shot 3D classification using rendered views. For 3D scene understanding, Michele \etal~\cite{Michele2021GenerativeZL} and Chen \etal~\cite{Chen2022ZeroshotPC} study zero-shot semantic segmentation with generative model and word embedding prototypes. Lu \etal~\cite{Lu2022OpenVocabulary3D} study zero-shot 3D object detection with pseudo-labels generated with a 2D classifier. These methods require point cloud labels on seen categories for supervised training, which still results in a model with close-set vocabularies. More recent works apply CLIP to 3D scenes for open-vocabulary tasks. However, they require human annotations for training, such as 2D pixel labels~\cite{Peng2022OpenScene3S} and 3D point labels~\cite{Ha2022SemanticAO}. PLA~\cite{Ding2022LanguagedrivenO3} utilizes an image-captioning model to generate captions for 3D content, bringing superior results on open-vocabulary segmentation by aligning 3D representations with text embeddings. Unlike existing works, we propose directly transferring CLIP's knowledge to 3D scene understanding models with no annotations. 
\vspace{-0.2cm}

\subsection{3D Representation Learning}
\vspace{-0.2cm}
Inspired by the success of self-supervised representation learning in 2D vision, 3D pre-training methods achieve better fine-tuning performance and efficiency in various down-stream tasks by leveraging contrastive learning~\cite{Hou2020ExploringD3,Xie2020PointContrastUP,Zhang2021SelfSupervisedPO,Liu2021TupleNCE,Chen20224DContrast}, masked auto-encoder~\cite{Yu2022PointBERT,Pang2022MaskedAF,Zhang2022PointM2AEMM,Dong2022AutoencodersAC}, or both~\cite{Qi2023ContrastWR}.

Beyond self-supervised pre-training, cross-modal learning methods propose to distill knowledge in images/text and pre-trained models to 3D representations~\cite{Afham2022CrossPointSC,Dong2022AutoencodersAC,Huang2023JointRL,Qi2023ContrastWR}. For scene understanding tasks, recent works enrich 3D point representations by utilizing 2D/text annotations~\cite{Yu2022DataE3,Zhang2022LanguageAssisted3F}, pixel-point alignments~\cite{Liu2021LearningF2,Sautier2022ImagetoLidarSD,Wang2022P2PTP,Li2022SimIPU}, neural rendering~\cite{Huang2022PonderPC} and pre-trained models~\cite{Qian2022Pix4PointIP,Rozenberszki2022LanguageGroundedI3,Xu2021Image2Point3P,Yao20223DPC,Zhang2023I2PMAE}. Our method can be regarded as an unsupervised cross-modal 3D representation learning method. We demonstrate that distilling CLIP's richly-structured vision knowledge to 3D models can benefit 3D scene understanding when limited annotations are available, outperforming previous self-supervised and cross-modal pre-training methods.
\vspace{-0.2cm}

\section{Method}
\vspace{-0.15cm}

This section describes the process of transferring CLIP's feature space to 3D scene understanding models. We first extract pixel-level CLIP features from the 3D scene's RGB views by modifying CLIP's inputs and forwarding process, introduced in Section~\ref{extract_feature}. We then get target point features from pixel features and train the 3D model by feature distillation, introduced in Section~\ref{3D_distillation}. 

\subsection{Extract Free Pixel-level CLIP Features}
\label{extract_feature}
\vspace{-0.2cm}

Although CLIP only aligns image-level global features with text embeddings, it should inherently encode local and dense semantics. As demonstrated in MaskCLIP~\cite{Zhou2021ExtractFD}, CLIP must divide image-level semantics into local segments, and properly align each segment's semantics with independent concepts in the text. MaskCLIP proposes to discard the global pooling layer and extract dense features from the final feature map with reformulated $1 \times 1$ convolutions.

However, adapting MaskCLIP to 3D scene contents brings poor dense prediction results, as shown in Section ~\ref{experiments}. On the one hand, CLIP's feature map has a much lower resolution than the input images (downsampled by $16^2$ in ViT/16~\cite{dosovitskiy2020vit} and $32^2$ in ResNet-50~\cite{He2016ResNet}). Although MaskCLIP is reasonably capable of recognizing salient semantics in the images, it struggles to segment the numerous objects in 3D scenes at different scales. On the other hand, the pixel features in CLIP's final feature map contain much global semantics regarding the entire image. They may contain multiple objects' semantics since each pixel's feature aggregates information from all other pixels in forwarding. However, ideally, we hope to extract object-level features of different objects in a 3D scene. We propose increasing the resolutions and extracting local features from CLIP to address the aforementioned problems, described as follows.

\textbf{Multi-scale region extraction.} Firstly, the input view is cropped at multi-scales to adapt to the recognition of objects of various sizes in 3D scenes, as shown in Figure \ref{main_fig} (a). For each cropped sample, we divide into many local regions. This effectively improves the feature resolution. Specifically, we divide the image sample into super-pixels with SLIC~\cite{Achanta2012SLICSC} as in~\cite{Sautier2022ImagetoLidarSD}. The super-pixel roughly covers an object or object's part, resulting in locally visually similar regions. After processing the input image, we extract an embedding vector for each super-pixel with modified CLIP's ViT~\cite{dosovitskiy2020vit} encoder, which is introduced below.

\textbf{Extracting local features from CLIP.} We hope the super-pixel feature aggregates information from local patches rather than from the entire image like the global classification token in ViT's encoder. This process is shown in Figure \ref{main_fig} (b). Given a cropped image sample $I_\text{c}$, we segment it into $N$ super-pixels: $I_\text{c} = \mathcal{S}_1 \cup \mathcal{S}_2 \cup \cdots \cup \mathcal{S}_N$, where $\mathcal{S}_i \cap \mathcal{S}_j=\varnothing,~ \forall~ i \neq j$. We then upsample the exact cropped image to CLIP's input size and obtain $I_\text{c}^{'}$. In ViT, $I_\text{c}^{'}$ is first reshaped into $M$ flattened patches: $I_\text{c}^{'} = \mathcal{P}_1 \cup \mathcal{P}_2 \cup \cdots \cup \mathcal{P}_M$, where $M=14^2$ in ViT-B/16. We then assign each patch in $I_\text{c}^{'}$ to a specific super-pixel in $I_\text{c}$ based on their spatial locations by interpolation, since the number of super-pixels $N$ is always fewer than the number of patches $M$. We denote the patch $\mathcal{P}_i$ being assigned to the super-pixel $\mathcal{S}_j$ as $\mathcal{P}_i \sim \mathcal{S}_j$.

To represent each super-pixel's feature, we add $N$ local classification tokens beyond the original global classification token in CLIP's forwarding process. The local tokens share some similarities with the group tokens in~\cite{Xu2022GroupViTSS}, but are not learnable and have different updating mechanism. The local tokens are initialized the same as the global one and are updated by the same self-attention mechanism and pre-trained weights in ViT. The only difference in updating the local tokens during inference is how attention scores are computed. Recall that in ViT's forwarding process, the attention score of the classification token is calculated as:
\begin{equation}
\begin{aligned}
    &\quad \ \ \ A_{\text{global}} = \sum_i \text{softmax} \left(\frac{q^{\text{g}} \cdot k_i^\mathrm{T}}{C} \right) v_i, \\
    q^{\text{g}} = &\text{Emb}_\text{q}(x^{\text{g}}), ~~ k_i = \text{Emb}_\text{k}(x_i), ~~ v_i = \text{Emb}_\text{v}(x_i),
\end{aligned}
\end{equation}
where $C$ is a constant scaling factor and $\text{Emb}(\cdot)$ denotes the linear layers encoding the query, key, and value embeddings. $x^{\text{g}}$ is the \textit{global} classification token and $x_i$ represents the input feature of patch $\mathcal{P}_i$.

In our method, the attention score of each local classification token is computed from local patches as:
\begin{equation}
\begin{aligned}
    & A_{\text{local, } j} = \sum_{i: \text{ } \mathcal{P}_i \sim \mathcal{S}_j} \text{softmax} \left(\frac{q^{\ell}_{j} \cdot k_i^\mathrm{T}}{C} \right) v_i,\\
    q^{\ell}_{j} = & \text{Emb}_\text{q}(x^{\ell}_{j}), ~k_i = \text{Emb}_\text{k}(x_i), ~ v_i = \text{Emb}_\text{v}(x_i),
\end{aligned}
\end{equation}
where $\mathcal{P}_i \sim \mathcal{S}_j$ means that patch $\mathcal{P}_i$ is assigned to the super-pixel $\mathcal{S}_j$. $x^{\ell}_{j}$ is the \textit{local} classification token of super-pixel $\mathcal{S}_j$ and $x_i$ represents the input feature of patch $\mathcal{P}_i$. 

Note that the original tokens in ViT are not affected during inference. By forwarding CLIP with additional tokens, we obtain a local feature for each super-pixel that is aligned with CLIP's feature space. We then apply the same local token feature to all pixels within a super-pixel to preserve object-level semantics. In this way, we obtain a feature map for each cropped image of the same size as CLIP's input.

\textbf{Multi-scale feature fusion.} After extracting local feature maps for all the cropped samples, we resize them back to the cropping sizes and stitch them back to the original input image, as shown in Figure \ref{main_fig} (c). Since one pixel may belong to different super-pixels from different cropped samples, we average all features as the final pixel feature. We extract pixel features for each 3D scene's RGB view. While the whole process is slow and difficult to apply to real-time inference, we only do the above process once for each view and use the pixel-level features as offline training targets.

\subsection{Feature Distillation with 2D Teacher}
\label{3D_distillation}
\vspace{-0.2cm}

To connect the pixel-level 2D features with 3D point features, we project each point in a 3D scene back to the RGB views following 3DMV~\cite{Dai20183DMVJ3}. Point-to-pixel projection is computed based on the camera pose, intrinsics, and the world-to-grid transformation matrix. Since we can obtain each projected pixel's depth value from the RGB-D images, we only keep the points that are in the correct depth ranges for further training. As some points will be associated with multiple pixels from different views, we compute the average of all 2D features as the final point feature $f_{\text{3D}} \in \mathbb{R}^C$, where the channel number $C$ is the same as CLIP's vision encoder. As a result, we obtain point-level features for each 3D scene denoted as $\mathcal{F}_{\text{target}} = \{\hat{f}_{\text{3D}, i}\}_{i=1}^{N_p}$, where $N_p$ is the number of points in the scene.

After extracting offline target point features for each scene, we train the 3D scene understanding model to learn from these targets by feature distillation. Denoting the learned point features for a scene as $\mathcal{F}_{\text{learn}} = \{f_{\text{3D}, i}\}_{i=1}^{N_p}$, the loss function of feature distillation is:

\vspace{-0.35cm}

\begin{equation}
\mathcal{L} = \frac{1}{N_p} \sum_i^{N_p} \mathcal{D}\left(f_{\text{3D}, i},~ \hat{f}_{\text{3D}, i}\right).
\vspace{-0.15cm}
\end{equation}

The distance $\mathcal{D}(\cdot, \cdot)$ is the negative cosine similarity:
\vspace{-0.3cm}

\begin{equation}
\mathcal{D}(f_1, f_2) = - \frac{f_1}{\| f_1 \|_{2}}  \cdot  \frac{f_2}{\| f_2 \|_{2}},
\vspace{-0.1cm}
\end{equation}
where $\| \cdot \|_{2}$ is $\mathcal{L}_2 \text{-norm}$. We use cosine distance because CLIP-driven classification relies on the cosine distance between vision and text embeddings.

The whole training process only requires the 3D dataset and the pre-trained CLIP vision encoder, without any form of supervision (\eg, 2D/3D annotations or vision-language grounding annotations). Since the learned point features are consistent with CLIP's feature space, the 3D model can perform open-vocabulary semantic segmentation and open-world reasoning once the feature distillation is finished.

\begin{table}[t]
\begin{center}
\footnotesize
\setlength\tabcolsep{5pt}
  \begin{tabular}{l|c|cc|cc}
    \toprule
     & \multirow{2}{*}{Backbone} & \multicolumn{2}{c}{ScanNet} & \multicolumn{2}{c}{S3DIS} \\
     &  & mIoU  & mAcc & mIoU  & mAcc \\
    \midrule
    \multicolumn{6}{l}{\textit{Inference by feature projection}} \\
    MaskCLIP-3D  & CLIP & 9.7 & 21.6 & - & - \\
    \rowcolor{linecolor} Target Feature (Ours) & CLIP & 27.6 & 47.7 & - & -  \\
    \midrule
    \rowcolor{linecolor} CLIP-FO3D (Ours) & Res-Unet & \textbf{30.2} & \textbf{49.1} & \textbf{22.3} & \textbf{32.8} \\
    \bottomrule
  \end{tabular}
  \end{center}
  \vspace{-10pt}
  \caption{Annotation-free semantic segmentation on standard ScanNet and S3DIS datasets. \textit{MaskCLIP-3D} and our \textit{Target Feature} use pixel features to infer points' semantics, which is slow and unsuitable for practical use. Our pixel-level target features capture significantly better semantics than MaskCLIP. CLIP-FO3D trained with feature distillation even outperforms its target.}
  \vspace{-0.4cm}
\label{table1}
\end{table}

\section{Experiments}
\label{experiments}

\subsection{Experimental Setup}
\vspace{-5pt}
\textbf{Dataset.} We train CLIP-FO3D on ScanNet's training set~\cite{Dai2017ScanNetR3}. We use the RGB-D images and the 3D scene meshes for training, and no labels are used. Specifically, we sub-sample RGB-D images from the raw ScanNet videos every ten frames for each scene. We evaluate our method on ScanNet~\cite{Dai2017ScanNetR3} and S3DIS~\cite{Armeni20163DSP} datasets. ScanNet's validation set and S3DIS's ``Area 5" are used for all the evaluation experiments. Notice that raw categories' names and the mapping to NYU40 label set are provided in the ScanNet dataset, which we use to enlarge the vocabulary.

\textbf{Implementation Details.} We use CLIP's ViT-B/16 encoder as the 2D backbone and use the corresponding text encoder for generating text embeddings. For all the experiments, we adopt MinkowskiNet16~\cite{Choy20194DSC} as 3D scene understanding backbone. We remove the final classifier and change the output feature dimension to 512 to match CLIP's feature dimension. We train CLIP-FO3D using SGD optimizer with a learning rate of 0.8 and a batch size of 4. The model is trained for 80K steps, and the learning rate is decreased by 0.99 for every 1,000 steps. The fine-tuning experiments on CLIP-FO3D are trained with a batch size of 4 for 40K steps. We set the initial learning rate of the pre-trained CLIP-FO3D networks to 0.001, with polynomial decay with power 0.9, because we find that a small learning rate leads to better results when fine-tuning CLIP's feature space. The initial learning rate of the classifier in data-efficient learning is set to 0.1 following the original setting in~\cite{Hou2020ExploringD3}. For zero-shot learning and data-efficient learning experiments, we follow the original benchmarks in~\cite{Chen2022ZeroshotPC} and~\cite{Hou2020ExploringD3}. More details can be found in Appendix \ref{sec:impl_detail}.

\begin{table}[t]
\begin{center}
\footnotesize
\setlength\tabcolsep{3.5pt}
  \begin{tabular}{l|cc|cc|cc}
    \toprule
     & \multicolumn{2}{c}{Head} & \multicolumn{2}{c}{Common} & \multicolumn{2}{c}{Tail} \\
     & mIoU  & mAcc & mIoU  & mAcc & mIoU  & mAcc \\
    \midrule
    \multicolumn{6}{l}{\textit{Inference by feature projection}} \\
    MaskCLIP-3D  & 19.8 & 33.1 & 13.3 & 26.8 & 7.5 & 15.3 \\
    \rowcolor{linecolor}Target Feature (Ours) & 37.1 & 63.4 & \textbf{39.0} & \textbf{55.4} & \textbf{40.6} & \textbf{55.3} \\
    \midrule
    \rowcolor{linecolor}CLIP-FO3D (Ours) & \textbf{44.3} & \textbf{65.9} & 37.6 & 50.6 & 26.5 & 36.4 \\
    \bottomrule
  \end{tabular}
  \end{center}
  \vspace{-10pt}
  \caption{Open-vocabulary semantic segmentation on ScanNet with extended labels from NYU label set. 
  Categories are divided into \textit{Head}, \textit{Common} and \textit{Tail} by point numbers. }
  \vspace{-0.4cm}
\label{table2}
\end{table}

\begin{table*}[t]
\begin{center}
\footnotesize
  \begin{tabular}{l|cccccccccc|c}
    \toprule
    & Monitor & Plant & Backpack & Printer & Trash can & Stove & Shoe & Blackboard & Radiator & Light & Mean \\
    \midrule 
    \multicolumn{6}{l}{\textit{Inference by feature projection}} \\
    MaskCLIP-3D & 35.8 & 28.1 & 7.1 & 8.9 & 5.2 & 26.3 & 3.6 & 20.8 & 14.6 & 17.7 & 16.8  \\
    \rowcolor{linecolor}Target Feature (Ours) & \textbf{83.7} & 72.8 & \textbf{57.0} & \textbf{30.1} & 63.2 & 69.0 & \textbf{21.9} & \textbf{82.9} & \textbf{54.8} & 40.8 & \textbf{57.6}\\
    \midrule 
    \rowcolor{linecolor}CLIP-FO3D (Ours) & 62.3 & \textbf{77.8} & 10.7 & 5.6 & \textbf{70.9} & \textbf{74.1} & 9.9 & 59.1 & 15.4 & \textbf{71.1} & 45.7 \\
    \bottomrule
  \end{tabular}
  \end{center}
  \vspace{-15pt}
  \caption{Annotation-free semantic segmentation on ScanNet \textbf{beyond NYU label set}. We show the results of the ten most frequent raw categories provided with the dataset beyond NYU labels. \textit{MaskCLIP-3D} and our \textit{Target Feature} use pixel features to infer points' semantics, which is slow and unsuitable for practical use. 
  }
\label{table3}
\vspace{-0.3cm}
\end{table*}

\begin{table*}[ht]
\begin{center}
\footnotesize
  \begin{tabular}{l|ccc|ccc|ccc|ccc|ccc}
    \toprule
    Setting & \multicolumn{3}{c}{Unseen-2} & \multicolumn{3}{c}{Unseen-4} & \multicolumn{3}{c}{Unseen-6} & \multicolumn{3}{c}{Unseen-8} & \multicolumn{3}{c}{Unseen-10} \\
    \midrule
    \multirow{2}{*}{Metric} & \multicolumn{2}{c}{mIoU} & \multirow{2}{*}{hIoU} & \multicolumn{2}{c}{mIoU} & \multirow{2}{*}{hIoU} & \multicolumn{2}{c}{mIoU} & \multirow{2}{*}{hIoU} & \multicolumn{2}{c}{mIoU} & \multirow{2}{*}{hIoU} & \multicolumn{2}{c}{mIoU} & \multirow{2}{*}{hIoU} \\
     & $\mathcal{S}$ & $\mathcal{U}$ & & $\mathcal{S}$ & $\mathcal{U}$ & & $\mathcal{S}$ & $\mathcal{U}$ & & $\mathcal{S}$ & $\mathcal{U}$ & & $\mathcal{S}$ & $\mathcal{U}$ \\
    \midrule
    3DGenZ~\cite{Michele2021GenerativeZL} & 33.4 & 12.8 & 18.5 & 32.8 & 7.7 & 12.5 & 31.2 & 4.8 & 8.3 & 29.7 & 2.1 & 3.9 & 30.1 & 1.4 & 2.7 \\
    TGP~\cite{Chen2022ZeroshotPC} & 58.6 & 51.6 & 54.9 & 57.9 & 34.1 & 42.9 & 55.2 & 15.4 & 24.1 & 51.6 & 12.1 & 19.6 & 52.5 & 9.5 & 16.1 \\
    MaskCLIP-3D & 67.3 & 38.7 & 49.1 & 64.4 & 31.0 & 41.9 & 62.9 & 24.7 & 35.5 & 61.8 & 19.1 & 29.2 & 62.3 & 13.4 & 22.1 \\
    \rowcolor{linecolor}CLIP-FO3D (Ours) & \textbf{70.6} & \textbf{60.3} & \textbf{65.0} & \textbf{69.5} & \textbf{54.8} & \textbf{61.2} & \textbf{67.3} & \textbf{50.8} & \textbf{57.9} & \textbf{65.8} & \textbf{46.7} & \textbf{54.6} & \textbf{67.7} & \textbf{40.7} & \textbf{50.8} \\
    \bottomrule
  \end{tabular}
  \end{center}
  \vspace{-15pt}
  \caption{Zero-shot semantic segmentation on ScanNet with different settings. ``Unseen-$i$"  indicates that there are $i$ classes that have no labels during training. $\mathcal{S}$ and $\mathcal{U}$ represent the performance of seen and unseen classes, respectively.}
\label{table4}
\vspace{-0.3cm}
\end{table*}

\begin{table}[ht]
\begin{center}
\footnotesize
  \begin{tabular}{l|ccc|ccc}
    \toprule
    Setting & \multicolumn{3}{c}{Unseen-4} & \multicolumn{3}{c}{Unseen-6}  \\
    \midrule
    \multirow{2}{*}{Metric} & \multicolumn{2}{c}{mIoU} & \multirow{2}{*}{hIoU} & \multicolumn{2}{c}{mIoU} & \multirow{2}{*}{hIoU}  \\
     & $\mathcal{S}$ & $\mathcal{U}$ & & $\mathcal{S}$ & $\mathcal{U}$ & \\
    \midrule
    3DGenZ~\cite{Michele2021GenerativeZL} & 53.1 & 7.3 & 12.9 & 28.3 & 6.9 & 11.1 \\
    TGP~\cite{Chen2022ZeroshotPC} & 60.4 &  20.6 & 30.7 & - & - & - \\
    \rowcolor{linecolor}CLIP-FO3D (Ours) & \textbf{64.8} & \textbf{26.1} & \textbf{37.2} & \textbf{60.8} & \textbf{29.3} & \textbf{39.5} \\
    \bottomrule
  \end{tabular}
  \end{center}
  \vspace{-15pt}
  \caption{Zero-shot semantic segmentation on S3DIS with different settings. ``Unseen-$i$" indicates there are $i$ classes that have no labels during training. $\mathcal{S}$ and $\mathcal{U}$ represent the performance of seen and unseen classes, respectively.}
\label{table5}
\vspace{-0.55cm}
\end{table}

\subsection{Annotation-free Semantic Segmentation}
\vspace{-5pt}
\textbf{Semantic segmentation on standard benchmarks.} We first present the results of annotation-free semantic segmentation on standard ScanNet and S3DIS benchmarks, where no supervision is provided during training. This is a rather challenging setting. After training CLIP-FO3D with feature distillation, we can take the text embeddings of each class name’s prompts as classification weights to perform semantic segmentation. We remove the ``other furniture" category in ScanNet and the ``clutter" category in S3DIS because they do not have specific semantics that can be classified with any text embeddings. 

The results are shown in Table \ref{table1}. The first two rows show the results of using multi-view pixel features to infer each point's semantics by feature projection (introduced in Section \ref{3D_distillation}). \textbf{\textit{MaskCLIP-3D}} is a baseline method that we use MaskCLIP to compute pixel features for RGB views as in~\cite{Zhou2021ExtractFD}. \textbf{\textit{Target Feature}} is our proposed method to extract free dense CLIP features from RGB views introduced in Section \ref{extract_feature}. Notice that one should forward the vision encoder for all RGB views and then align 3D points with pixels with camera pose, intrinsics and transformation matrix. This inference process by feature projection is very slow and not applicable for practical use. In contrast, CLIP-FO3D is convenient for processing new 3D data. However, the results can still be compared and give some inspiration. 

It is observed that our Target Features extract more reasonable semantics than MaskCLIP-3D, indicating that the proposed method is indispensable in extracting dense CLIP features for 3D contents. Moreover, our CLIP-FO3D achieves remarkable segmentation results on ScanNet and S3DIS in this challenging setting (notice that four categories in S3DIS never appear during the training on ScanNet). CLIP-FO3D even outperforms its target features in ScanNet, indicating that some misleading target features can be corrected during feature distillation.

\textbf{Semantic segmentation with open vocabularies.} The standard ScanNet benchmark only contains a small vocabulary of 20 classes. To examine the open-vocabulary capability of CLIP-FO3D inherited from CLIP, we first extend the original vocabulary size with the NYU-40 label set. We remove the NYU-40 labels that do not have specific semantics (\eg, ``other structure", ``other furniture", ``other prop") and evenly divide all the rest categories into \textit{Head}, \textit{Common} and \textit{Tail} based on the sample numbers of each category. We show the semantic segmentation results on the three category sets in Table \ref{table2}. Our Target Features outperform the baseline MaskCLIP-3D by a large margin. Since categories in \textit{Common} and \textit{Tail} set usually contain objects with smaller sizes (\eg, bag, box, pillow, book), MaskCLIP-3D struggles to extract their semantics due to the limited feature resolution. While our method even extracts higher quality features for \textit{Tail} categories than \textit{Head}, thanks to the multi-scale inputs and local super-pixel features. CLIP-FO3D also achieves great results in all categories, while it does not perform as well as our Target Features for \textit{Common} and \textit{Tail} categories, due to the limited number of examples in the training set. However, CLIP-FO3D can be easily used for indoor open-vocabulary scene understanding applications.

We further show the semantic segmentation results of more categories beyond the NYU-40 label set: the ten most frequent raw categories provided with the ScanNet dataset that do not belong to NYU-40 categories are used in Table \ref{table3}. Our Target Features and CLIP-FO3D perform well on these long-tailed categories, showing great open-vocabulary scene understanding properties.

\begin{figure*}[ht!]
\begin{center}
\centerline{\includegraphics[width=1.9\columnwidth]{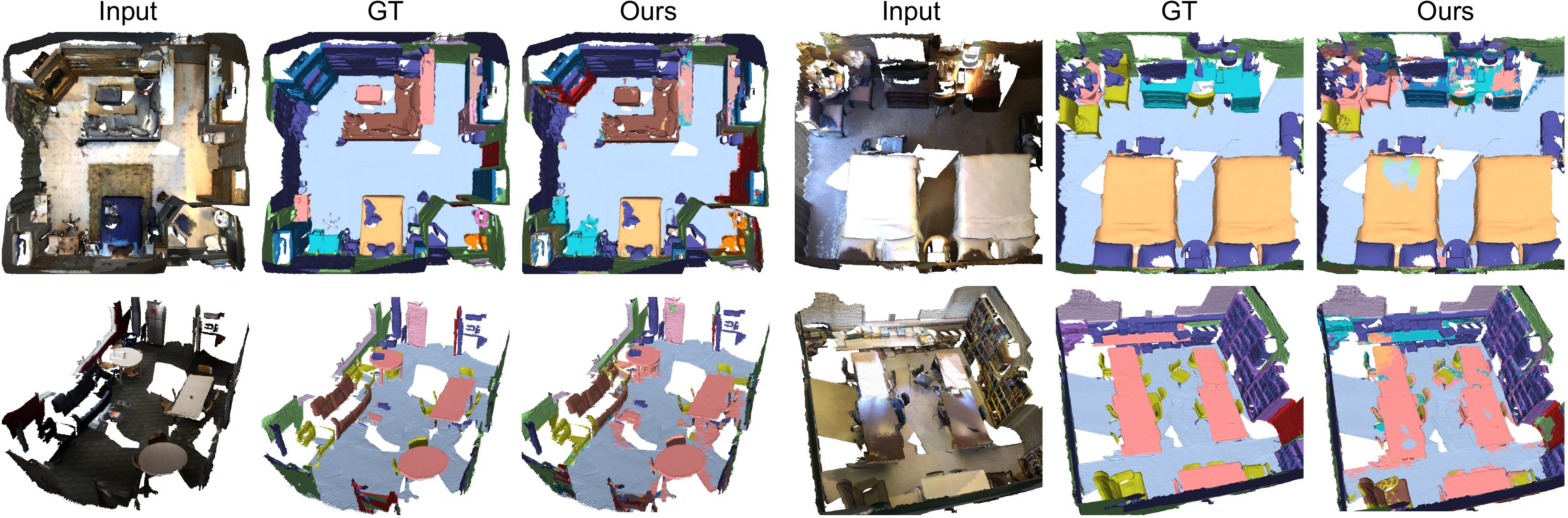}}
\caption{Qualitative results of annotation-free semantic segmentation on standard ScanNet benchmark.}
\label{visual}
\end{center}
\vspace{-1.0cm}
\end{figure*}

\begin{figure}[t]
\begin{center}
\centerline{\includegraphics[width=0.85\columnwidth]{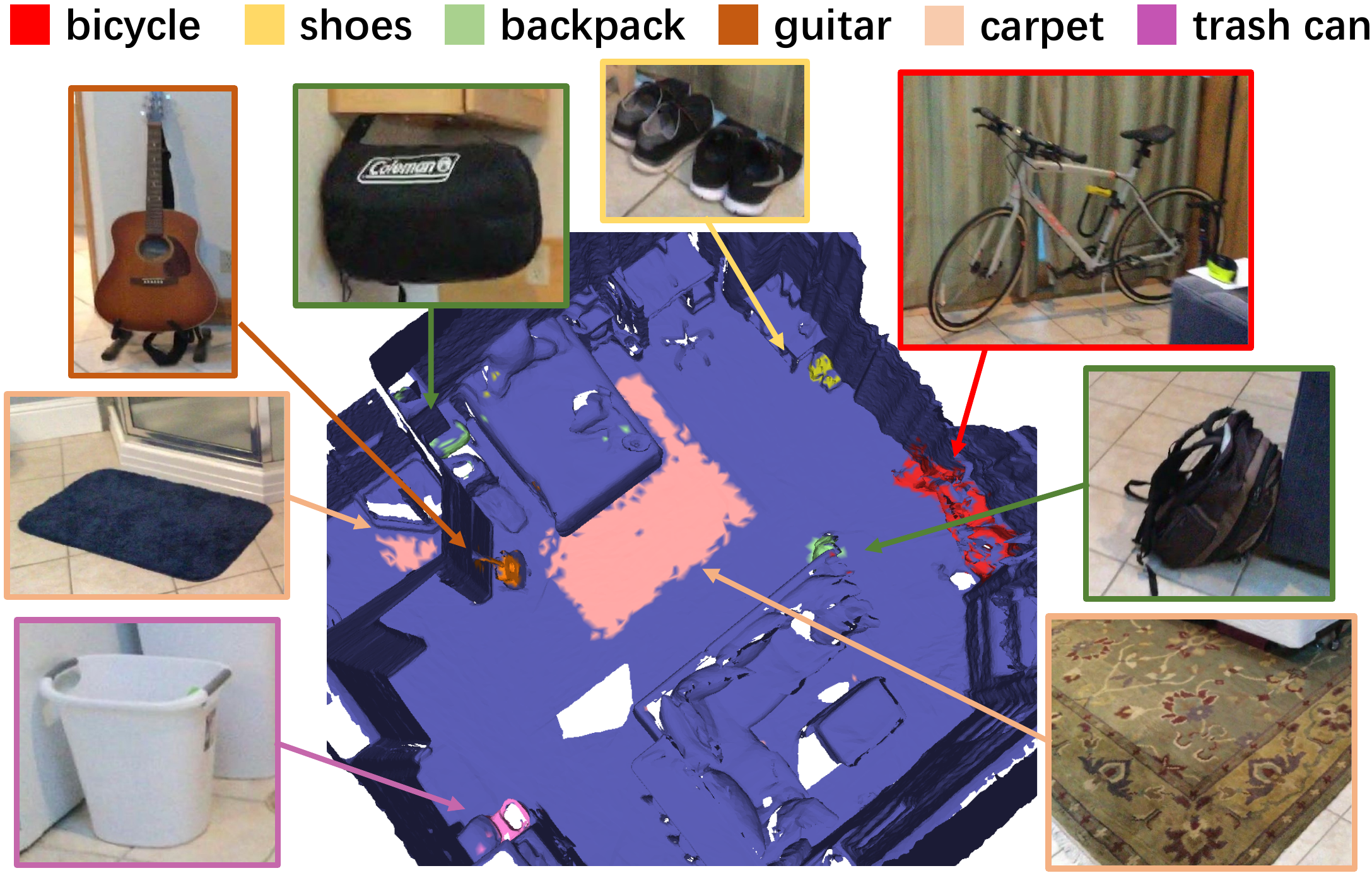}}
\caption{Qualitative results of annotation-free semantic segmentation for long-tailed categories on ScanNet.}
\label{longtail_visual}
\end{center}
\vspace{-1.2cm}
\end{figure}

\textbf{Qualitative Results.} We show the visualization of annotation-free semantic segmentation results on standard ScanNet benchmark and long-tailed categories. As shown in Figure \ref{visual}, decent segmentation masks are obtained compared to the ground truth. Our method differs from ground-truth results for some objects with ambiguous semantics, such as \texttt{table} and \texttt{desk}, \texttt{cabinet} and its \texttt{door}, \texttt{chair} and \texttt{sofa}. In Figure \ref{longtail_visual}, our method successfully segments several long-tailed categories that are not annotated in traditional benchmarks, demonstrating our model's excellent open vocabulary capability.

\subsection{Zero-shot Semantic Segmentation}
\vspace{-3pt}
Zero-shot semantic segmentation methods train the 3D scene understanding model only with the labels on a subset of classes (seen classes) and evaluate both seen classes and unseen classes. All the existing methods in 3D scene understanding use transductive settings, in which the unlabeled points are accessible during training. CLIP-FO3D can be applied to zero-shot semantic segmentation with minor effort. Specifically, CLIP-FO3D can be used to generate pseudo-labels for the unlabeled points.

We compare our method with state-of-the-art zero-shot semantic segmentation methods on ScanNet and S3DIS following the benchmarks in~\cite{Chen2022ZeroshotPC}. We also build a new benchmark following~\cite{Ding2022LanguagedrivenO3}, where six classes are chosen as unseen classes in S3DIS. We use the metric of mean intersection over union (mIoU) for both seen classes ($\mathcal{S}$) and unseen classes ($\mathcal{U}$), and use the harmonic mean IoU (hIoU) to demonstrate the overall performance of zero-shot learning as in~\cite{Bucher2019ZeroShotSS,Xian2017ZeroShotLC}.

As shown in Table \ref{table4}, our method outperforms the previous state-of-the-art method~\cite{Chen2022ZeroshotPC} by 10.1\%, 18.3\%, 33.8\%, 35.0\% and 34.7\% on hIoU metric when there are 2, 4, 6, 8 and 10 unseen classes during training. Our CLIP-FO3D also outperforms the method of pseudo-labeling with MaskCLIP-3D by large margins. It is observed that our method is more effective than the baselines when there are more unseen classes, demonstrating  superior zero-shot capability. The results on S3DIS with two settings in Table \ref{table5} reflect a similar phenomenon, although S3DIS's data is not accessible during the training of CLIP-FO3D.

\begin{table}[t]
\begin{center}
\footnotesize
\setlength\tabcolsep{3.8pt}
  \begin{tabular}{l|c|cccc}
    \toprule
    \multirow{2}{*}{Method} & \multirow{2}{*}{Sup.}  & \multicolumn{4}{c}{Percentage of labeled scene} \\
     &  & 1\% & 5\% & 10\% & 20\%  \\
    \midrule
    Scratch & - & 26.8 & 46.9 & 58.4 & 62.9 \\
    CSC~\cite{Hou2020ExploringD3} & \xmark & 29.7 & 48.8 & 59.9 & 64.7  \\
    LangGround~\cite{Rozenberszki2022LanguageGroundedI3} & \cmark & 31.9 & 49.3 & 60.8 & 65.3 \\
    \rowcolor{linecolor}CLIP-FO3D (Ours) & \xmark & \textbf{36.1} \tiny\textcolor{sun-orange}{\textbf{($\uparrow$9.3)}} & \textbf{51.7} \tiny\textcolor{sun-orange}{\textbf{($\uparrow$4.8)}} & \textbf{63.1} \tiny\textcolor{sun-orange}{\textbf{($\uparrow$4.7)}} & \textbf{66.8} \tiny\textcolor{sun-orange}{\textbf{($\uparrow$3.9)}} \\
    \bottomrule
  \end{tabular}
  \end{center}
  \vspace{-10pt}
  \caption{Data-efficient semantic segmentation on ScanNet with \textbf{limited scene reconstructions}. ``Sup." indicates whether semantic labels are used during pre-training.}
  \vspace{-15pt}
\label{table6}
\end{table}

\begin{table}[t]
\begin{center}
\footnotesize
\setlength\tabcolsep{3.6pt}
  \begin{tabular}{l|c|cccc}
    \toprule
    \multirow{2}{*}{Method} & \multirow{2}{*}{Sup.}  & \multicolumn{4}{c}{Number of labeled points} \\
    &  & 20 & 50 & 100 & 200  \\
    \midrule
    Scratch & - & 46.3 & 58.3 & 62.8 & 65.4 \\
    CSC~\cite{Hou2020ExploringD3} & \xmark & 54.0 & 60.7 & 65.6 & 68.3  \\
    LangGround~\cite{Rozenberszki2022LanguageGroundedI3} & \cmark & 55.1 & 62.4 & 66.0 & 68.2 \\
    \rowcolor{linecolor}CLIP-FO3D (Ours) & \xmark & \textbf{57.6} \tiny\textcolor{sun-orange}{\textbf{($\uparrow$11.3)}} & \textbf{64.3} \tiny\textcolor{sun-orange}{\textbf{($\uparrow$6.0)}} & \textbf{68.2} \tiny\textcolor{sun-orange}{\textbf{($\uparrow$5.4)}} & \textbf{69.5} \tiny\textcolor{sun-orange}{\textbf{($\uparrow$4.1)}} \\
    \bottomrule
  \end{tabular}
  \end{center}
  \vspace{-10pt}
  \caption{Data-efficient semantic segmentation on ScanNet with \textbf{limited point annotations}. ``Sup." indicates whether points' semantic labels are used during pre-training.}
  \vspace{-6pt}
\label{table7}
\end{table}

\subsection{Data-efficient 3D Scene Understanding}
As the collection and annotation of 3D point cloud data are very laborious, data-efficient learning methods have been proposed to train a better 3D model when training data or labels are scarce. Existing works have explored self-supervised~\cite{Hou2020ExploringD3} or cross-modal~\cite{Rozenberszki2022LanguageGroundedI3} pre-training methods for data-efficient fine-tuning. CLIP-FO3D can also be considered an unsupervised cross-modal pre-training method to benefit data efficiency. Unlike existing works, our method leverages the richly-structured feature space inherited from CLIP to improve data-efficient fine-tuning results.

\begin{table}[t]
\begin{center}
\footnotesize
  \begin{tabular}{l|cc|ccc}
    \toprule
     \multirow{2}{*}{Method} & \multicolumn{2}{c}{Annotation-free} & \multicolumn{3}{c}{Zero-shot (Unseen-4)}  \\
     & mIoU & mAcc & $\mathcal{S}$ & $\mathcal{U}$ & hIoU \\
    \midrule
    w/o Multi-scale & 17.1 & 33.8 & 68.1 & 43.3 & 52.9 \\
    w/o Local Feature & 21.6 & 39.3 & 68.3 & 48.5 & 56.7 \\      
    \rowcolor{linecolor}CLIP-FO3D (Ours) & \textbf{30.2}  & \textbf{49.1} & \textbf{69.5} & \textbf{54.8} & \textbf{61.2} \\
    \bottomrule
  \end{tabular}
  \end{center}
  \vspace{-3pt}
  \caption{Ablation study on ScanNet for two benchmarks. ``Multi-scale" represents cropping input images at multi-scales when extracting pixel-level features with CLIP. ``Local Feature" represents using local classification tokens to extract super-pixel features.}
  \vspace{-8pt}
\label{table8}
\end{table}

\begin{figure*}[t]
\begin{center}
\centerline{\includegraphics[width=1.95\columnwidth]{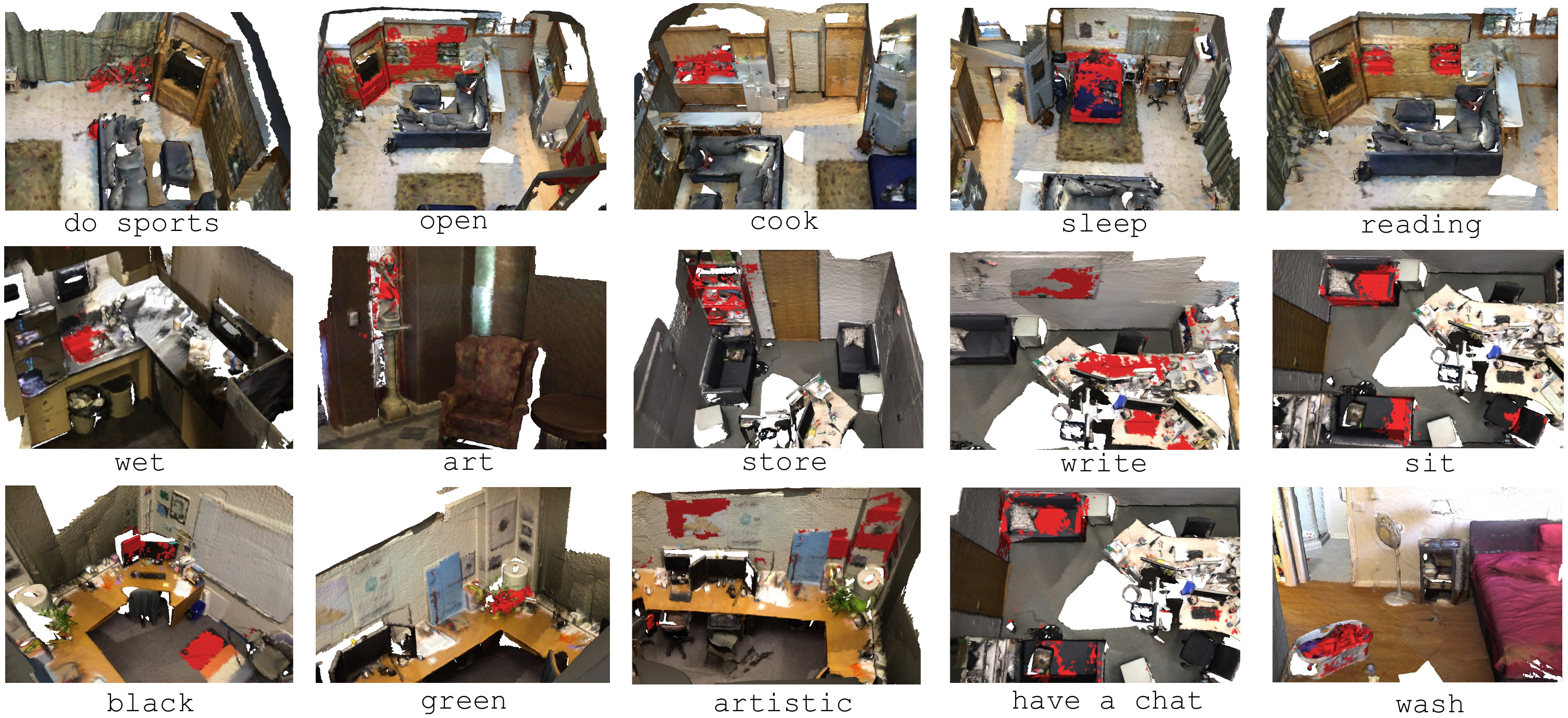}}
\caption{Open-world 3D scene understanding on ScanNet. Texts are queries and the most relevant areas are marked in \textcolor{my_red}{red}.}
\label{open_world}
\end{center}
\vspace{-1.0cm}
\end{figure*}

We follow the official data-efficient learning benchmarks~\cite{Hou2020ExploringD3}: Limited Reconstructions (only a few labeled scenes are used for training) and Limited Annotations (only a few points are labeled in each scene). ``Scratch" denotes the training from scratch baseline. The results of limited scene reconstructions are shown in Table \ref{table6}, when only 1\%, 5\%, 10\%, and 20\% scenes are used during training. Our method achieves mIoU improvements of 9.3\%, 4.8\%, 4.7\%, and 3.9\% over training from scratch, and achieves superior results compared to the previous state-of-the-art supervised pre-training method. The results of limited point annotations are shown in Table \ref{table7}, when 20, 50, 100, and 200 annotated points are randomly sampled for each scene. Our method achieves mIoU improvements of 11.3\%, 6.0\%, 5.4\%, and 4.1\% over training from scratch, and outperforms previous state-of-the-art supervised pre-training methods.

\section{Discussion}
\label{disc}

\textbf{Ablation study.} Table \ref{table8} shows the ablation study on the proposed methods when extracting pixel features from CLIP. ``Multi-scale" represents increasing the feature resolution with multi-scale inputs, and ``Local Feature" represents extracting super-pixel features using local classification tokens. We show the results of annotation-free and zero-shot semantic segmentation (with four unseen classes) on ScanNet. It is observed that both techniques are crucial for transferring CLIP's feature space. While each single method outperforms the MaskCLIP-3D baseline, increasing the feature resolution with multi-scale inputs brings more significant improvements than extracting local features.

\textbf{Open-world 3D scene understanding.} Since CLIP is trained with massive image-text pairs mined from the Internet, it inherently encodes rich real-world knowledge which can guide open-world applications such as visual navigation and embodied AI~\cite{Gadre2022CLIPOW,Khandelwal2021SimpleBE}. Since we directly transfer CLIP's feature space to 3D models, CLIP-FO3D has the potential to accomplish open-world 3D scene understanding. 

Inspired by the qualitative results in~\cite{Lddecke2021ImageSU} and~\cite{Peng2022OpenScene3S}, we use text embeddings to query open-world semantics for CLIP-FO3D's scene representations. Given a text description, we extract its feature with CLIP's text encoder, calculate its similarity with point features, and then threshold to produce a 3D mask. The visualization results are shown in Figure \ref{open_world}. Our model successfully finds the location most relevant to the text descriptions' actual meaning. For example, given ``\texttt{write}", CLIP-FO3D finds the whiteboard and desk on which one can write. Our model also understands other affordance, activity, color, and function words. These results demonstrate that our 3D scene representations encode rich and well-structured knowledge about the real world.

Trained with no human annotations, CLIP-FO3D successfully preserves CLIP's open-world properties, while the model trained with human annotations can only recognize object concepts. Our method broadens the applications for 3D scene understanding. For example, in conjunction with the language foundation models (\eg, BERT~\cite{Devlin2019BERTPO}, GPT-3~\cite{Brown2020LanguageMA}), CLIP-FO3D may enrich the functionality of robots by connecting 3D vision and language modalities.

\section{Conclusion}
\vspace{-0.25cm}
This paper proposes directly transferring CLIP's feature space to 3D scene understanding models without any supervision. We first extract pixel-level features from CLIP for 3D scene's RGB views. Then we adopt the feature projection scheme to get the target point features and train a 3D model with feature distillation. CLIP-FO3D achieves remarkable annotation-free semantic segmentation results on standard benchmarks as well as open-vocabulary concepts. Our model also outperforms previous state-of-the-art methods in zero-shot and data-efficient learning tasks. Most importantly, our model successfully inherits CLIP’s open-world properties, allowing 3D scene understanding models to encode open-world knowledge beyond object concepts.
\vspace{-0.25cm}

{\small
\bibliographystyle{ieee_fullname}
\bibliography{egbib}
}


\twocolumn[{%
\renewcommand\twocolumn[1][]{#1}%
\begin{center}
    \centering
    \captionsetup{type=figure}
    \includegraphics[width=1.0\linewidth]{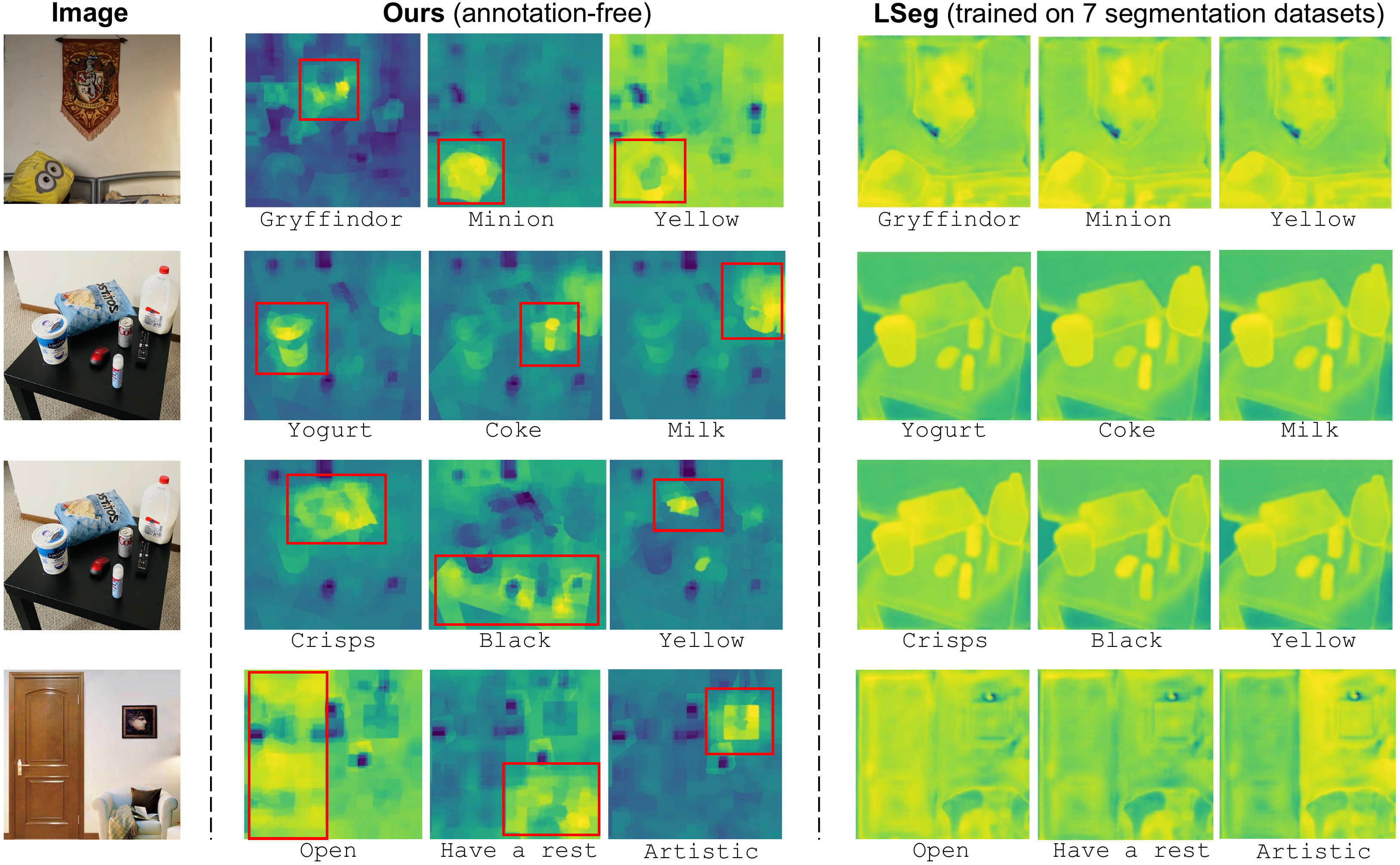}
    \captionof{figure}{Visualization of pixel-text similarities of our method and LSeg~\cite{Li2022LanguagedrivenSS}. Colors from blue to yellow indicate the similarity from low to high. We show the text queries regarding \textbf{long-tailed concepts} and \textbf{open-world knowledge} (\eg, color and affordance). Our method successfully localizes these queries thanks to the strong open-world properties of CLIP and our annotation-free setting. LSeg produces non-discriminative results since the required knowledge is forgotten during supervised training and not encoded in LSeg's feature space.}\label{Lseg}
\end{center}%
}]

\appendix 
\section{Advantages of Annotation-free Training}
Our contribution is directly transferring CLIP feature space to 3D representations without any form of annotation. We demonstrate that this annotation-free training preserves CLIP's rich information to the maximum. In contrast, fine-tuning CLIP with human annotations restricts the feature space. And the open-world knowledge encoded in CLIP is \textbf{forgotten} during fine-tuning as demonstrated in~\cite{Jatavallabhula2023ConceptFusionOM}.

To further demonstrate the advantages of annotation-free training, we compare our unsupervised pixel-level feature extraction with a supervised 2D segmentation method, LSeg~\cite{Li2022LanguagedrivenSS}. Lseg aligns pixel embeddings to the CLIP text embedding of the corresponding semantic class. It is trained on 7 segmentation datasets, and most importantly, uses the original label sets provided by these datasets without relabeling (details can be found in Section 5.2 of Lseg's paper~\cite{Li2022LanguagedrivenSS}). Large amount of human annotations are required, and the supervision covers hundreds of categories in LSeg.

Although our annotation-free method does not perform well compared with LSeg on some common categories from existing benchmarks, we show that our method has outstanding advantages of recognizing \textbf{long-tailed categories} and \textbf{open-world knowledge} (\eg, color, affordance) directly inherited from CLIP. Although Lseg is trained on large-scale annotated object concepts, its feature space is restricted to these concepts and can not be generalized to other open-world knowledge.

Visualization results of capturing long-tailed concepts and open-world knowledge are shown in Figure \ref{Lseg}. We use the ViT-B/16-based Lseg model that is pre-trained on 7 datasets provided by its official code. The similarity of the pixels and text queries are used for visualization (normalized to $[0,1]$). The second picture in Figure \ref{Lseg} comes from the paper of a concurrent work~\cite{Jatavallabhula2023ConceptFusionOM}, which proposes mapping language, vision and audio inputs to the same feature space. We show the results of long-tailed categories (\eg, \texttt{yogurt}, \texttt{milk}, \texttt{crisps}) and some proper names (\eg, \texttt{Gryffindor}, name of a house in the Harry Potter films; \texttt{Minion}, classic cartoon character). These categories are not included in LSeg's pre-training datasets, so it produces non-discriminative results, while our method successfully highlights the target regions. Besides, we show the results of open-world text queries, such as color and affordance words. It is observed that our model indeed encodes color information into pixel features (\eg, \texttt{black}, \texttt{yellow}) and also understands objects' affordance (\eg, \texttt{open}, \texttt{rest} and \texttt{artistic}). Since these open-world text queries are irrelevant to the object concepts in pre-training datasets, LSeg does not encode this knowledge into its feature space.

A concurrent work~\cite{Peng2022OpenScene3S} utilizes a pre-trained LSeg model to perform zero-shot 3D semantic segmentation. Given that LSeg is trained on a large amount of human annotations regarding hundreds of raw categories, and pixel-to-point alignments can also be obtained from camera poses and intrinsics in~\cite{Peng2022OpenScene3S}, we tend to consider their method \textit{not really} zero-shot learning for 2D annotations. In contrast, CLIP-FO3D is entirely annotation-free.

\section{Implementation Details}\label{sec:impl_detail}
\textbf{More details of the training process.} When extracting free pixel-level CLIP features, we introduce multi-scale region extraction and super-pixel partition in Section \ref{extract_feature}. For multi-scale region extraction, we use the cropping size of $1, \frac{1}{2}$, and $\frac{1}{4}$ of the input view's width/height. And the cropping window slides with the stride of $\frac{1}{2}$ cropping size. For super-pixel partition, we divide each image sample into $50$ super-pixels with SLIC~\cite{Achanta2012SLICSC}. We use a smaller number of super-pixel compared with SLidR~\cite{Sautier2022ImagetoLidarSD} since we already improved the resolution with multi-scale cropping. 
Through super-pixel partition, we extract locally visually similar regions which represent objects or object parts. By extracting a local feature for each super-pixel, we intend to preserve object-level CLIP semantics.

\textbf{More details of the ablation study.} Here, we introduce the implementation details of the ablation study in Section \ref{disc}. In the ``w/o Multi-scale" setting, we discard the multi-scale region extraction and the corresponding multi-scale feature fusion. We only extract local super-pixel features for the whole input view. In the ``w/o Local Features" setting, multi-scale cropping and feature fusion are kept the same as in the main method, and the CLIP local feature extraction process is replaced with MaskCLIP~\cite{Zhou2021ExtractFD}. 

\textbf{Prompt engineering.} To classify points' semantics, we take the text embeddings of each class name's prompts as classification weights. We use the $80$ hand-craft prompts as in MaskCLIP. In practice, we extract CLIP embeddings for all prompts and average them to obtain a single text embedding for each class.

\section{Dataset Partitioning}

\textbf{Semantic segmentation with open vocabularies.} To examine the open-vocabulary properties of CLIP-FO3D, we extend the
original vocabulary size in ScanNet~\cite{Dai2017ScanNetR3} benchmark with the NYU-40 label set. We remove the NYU-40 labels that do not have specific semantics (\eg, ``other structure", ``other furniture", ``other prop") and evenly divide all the rest categories into \textit{Head}, \textit{Common} and \textit{Tail}. \textbf{Head} classes contain wall, floor, cabinet, bed, chair, bathtub, table, door, toilet, bookshelf, curtain, and ceiling. \textbf{Common} classes contain sofa, counter, desk, dresser, refrigerator, shelves, shower curtain, night stand, window, picture, sink and floor mat. \textbf{Tail} classes contain blinds, mirror, clothes, pillow, book, box, whiteboard, lamp, towel, bag, person, and television.

\textbf{Zero-shot learning} We follow the original benchmarks in~\cite{Chen2022ZeroshotPC} for zero-shot semantic segmentation. In ScanNet, we conduct experiments with a different number of unseen classes, including the 2-sofa/desk, 4-bookshelf/toilet, 6-bathtub/bed, 8-curtain/window, 10-door/counter. In S3DIS~\cite{Armeni20163DSP}, beam, column, window and sofa are unseen classes. We also conduct experiments on a new benchmark with six unseen classes in S3DIS, where board, door, floor, sofa, table, window are chosen as unseen classes. This benchmark is first used in PLA~\cite{Ding2022LanguagedrivenO3}, which aligns coarse-to-fine 3D scene representations with paired text embeddings and achieves promising zero-shot learning results. Different with this work, our method does not rely on pre-trained image-captioning model and text data.

\section{More Qualitative Results} 
\begin{figure*}[ht]
\begin{center}
\centerline{\includegraphics[width=1.8\columnwidth]{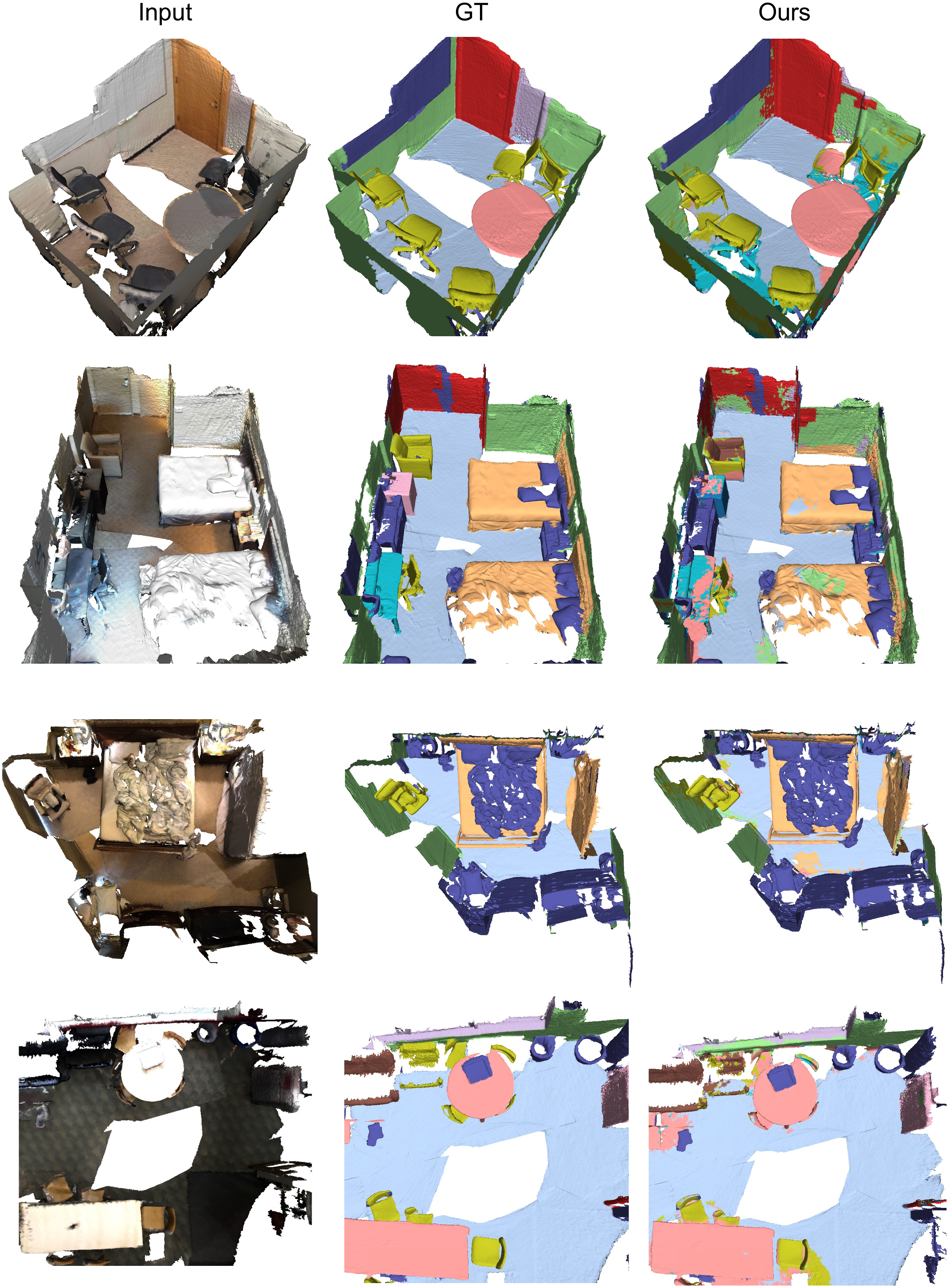}}
\caption{Visualization of annotation-free semantic segmentation on ScanNet.}
\label{segmentation}
\end{center}
\end{figure*}

\begin{figure*}[ht]
\begin{center}
\centerline{\includegraphics[width=1.8\columnwidth]{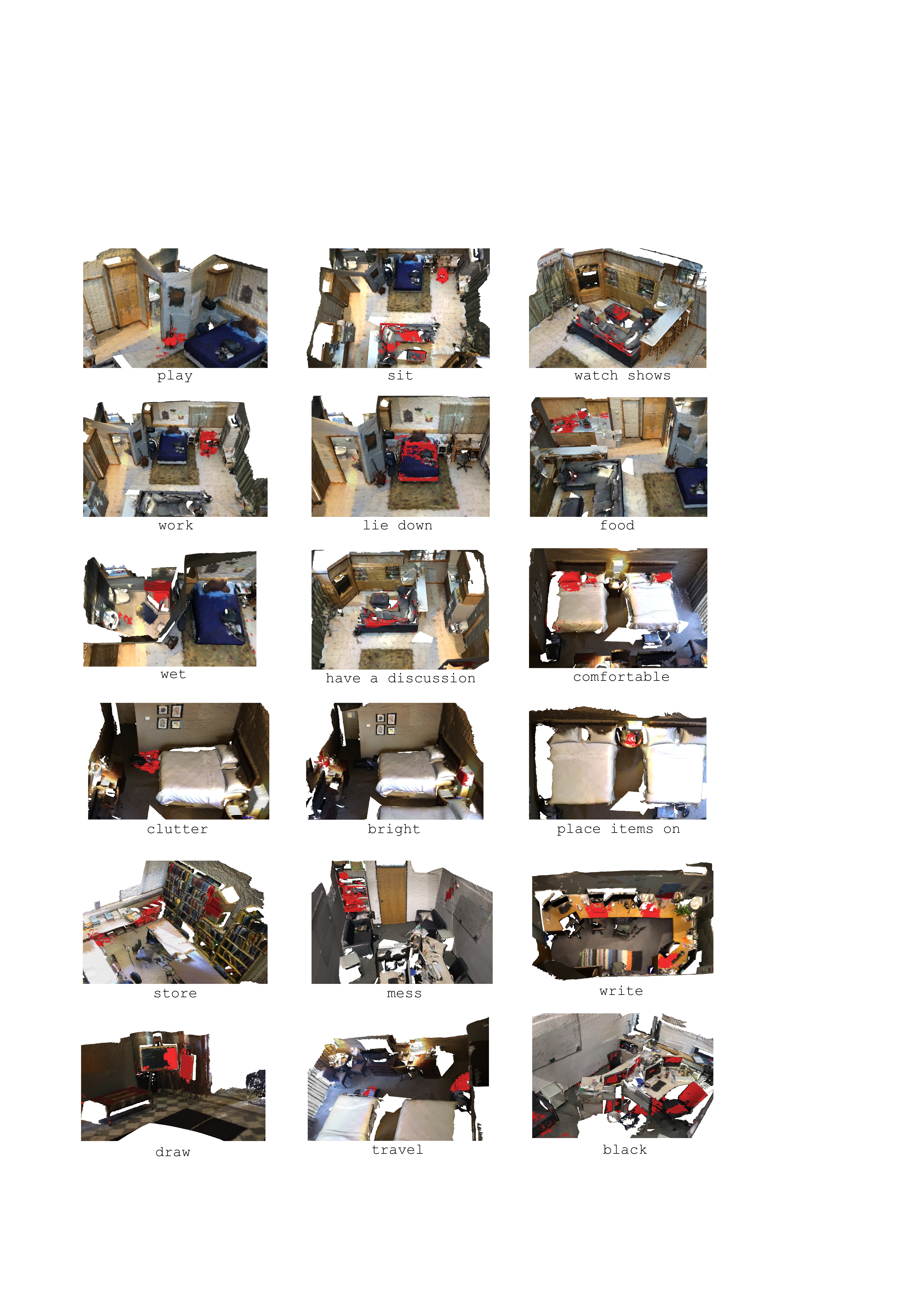}}
\caption{Open-world 3D scene understanding on ScanNet. Texts are queries and the most relevant areas are marked in \textcolor{my_red}{red}.}
\label{open_world_appendix}
\end{center}
\end{figure*}

We show additional qualitative results of annotation-free semantic segmentation in Figure \ref{segmentation} and open-world 3D scene understanding in Figure \ref{open_world_appendix}.


\end{document}